\documentclass[sigconf, nonacm]{acmart}
\renewcommand\footnotetextcopyrightpermission[1]{} 
\usepackage{multirow}
\AtBeginDocument{%
  \providecommand\BibTeX{{%
    \normalfont B\kern-0.5em{\scshape i\kern-0.25em b}\kern-0.8em\TeX}}}

\setcopyright{acmcopyright}
\copyrightyear{2020}
\acmYear{2020}
\acmDOI{}

\begin{document}

\title{Adaptive Adversarial Logits Pairing}

\author{Shangxi Wu}
\email{kirinng0709@gmail.com}
\affiliation{%
  \institution{Beijing Jiaotong University}
  \city{Beijing}
  \state{China}
}

\author{Jitao Sang}
\affiliation{%
  \institution{Beijing Jiaotong University}
  \city{Beijing}
  \state{China}
}

\author{Kaiyuan Xu}
\affiliation{%
  \institution{Beijing Jiaotong University}
  \city{Beijing}
  \state{China}
}

\author{Guanhua Zheng}
\affiliation{%
  \institution{University of Science and Technology of China}
  \city{Beijing}
  \state{China}
}

\author{Changsheng Xu}
\affiliation{%
  \institution{Institute of Automation, Chinese Academy of Sciences}
  \city{Beijing}
  \state{China}
}

\begin{abstract}
  Adversarial examples provide an opportunity as well as impose a challenge for understanding image classification systems. Based on the analysis of the adversarial training solution—Adversarial Logits Pairing (ALP), we observed in this work that: (1) The inference of adversarially robust model tends to rely on fewer high-contribution features compared with vulnerable ones. (2) The training target of ALP doesn't fit well to a noticeable part of samples, where the logits pairing loss is overemphasized and obstructs minimizing the classification loss. Motivated by these observations, we design an Adaptive Adversarial Logits Pairing (AALP) solution by modifying the training process and training target of ALP. Specifically, AALP consists of an adaptive feature optimization module with Guided Dropout to systematically pursue fewer high-contribution features, and an adaptive sample weighting module by setting sample-specific training weights to balance between logits pairing loss and classification loss. The proposed AALP solution demonstrates superior defense performance on multiple datasets with extensive experiments.
\end{abstract}




\maketitle

\section{Introduction}
Computer Vision discipline, such as Image Classification, has made a major breakthrough in recent years. However, an adversarial attack can easily deceive these models~\cite{szegedy2013intriguing, kurakin2016adversarial, Seyed-Mohsen2016Deepfool, Ian2017Explaining, nicholas2017towards}. Take Image Classification as an example, Adversarial attack is a technic that adds subtle perturbations which are hard for humans to detect in order to change output result dramatically. Recently, the adversarial attack is not limited in Image Classification problems but has been expanded to Object Detection, Face Recognition, Voice Recognition and Text Recognition ~\cite{advobdetection, adv_face, adv_voice, adv_text}. Hence, it is highly needed to design such models to defend adversarial samples ~\cite{8649865, 8576563}.

There are three main methods for defending existing adversarial samples~\cite{8884184, 8970483, Fangzhouy2017Defense, madry2017towards, papernot2016distillation, guo2017countering, dhillon2018stochastic, song2017pixeldefend, tramer2017ensemble, liu2018feature, liao2017defense, shen2017ape}: 
(1) Gradient masking: Hiding gradient information of images through nondifferentiable transformation methods in order to invalidate adversarial attack~\cite{shen2017ape}.
(2) Adversarial samples detecting: Preventing samples that have been attacked from being imported into the model by detecting features of adversarial samples~\cite{detecting}.
(3) Adversarial training: Continuously adding adversarial samples into the training sets in order to get robust model parameters~\cite{madry2017towards}.
Athalye \emph{et al.} ~\cite{athalye2018obfuscated} has released the limitations of the gradient masking methods, finding that they can still be attacked successfully by gradient simulation. In addition, adversarial samples detecting also has limitations of not being able to solve the problem of adversarial attack, which makes no contribution to the correction of the model’s fragileness.
In contrast with Gradient masking and Adversarial samples detecting, adversarial training is the most effective method which can improve the robustness of models.

Adversarial training was a simple training framework that aims at minimizing the cross-entropy of the original sample and adversarial sample at the same time. Under the training framework of adversarial training, Adversarial Logits Pairing (ALP)~\cite{alp}, which is a more strict adversarial training constraint, acts more effectively. 
ALP can not only constraint the cross-entropy of the original sample and adversarial sample, but can also require models to keep similar outputs between original samples and adversarial samples on logits activation.  
The current SOTA method TRADES~\cite{ZhangYJXGJ19} also uses the same idea as the ALP method, constraining the original sample and the adversarial sample to have similar representations.

When analyzing the result of models trained by ALP, we found that ALP loss had led to different effects on different samples,
as in Fig.~\ref{fig_2_example} shows that models trained by ALP present high conference in successfully defended samples but a low conference in those failed ones. 
By analyzing different models with different robustness, as shown in Fig.~\ref{feature_bf}, we found that according to the improvement of robustness, models would show fewer high-contribution features. 
Hence, we think ALP still has large room to be promoted. 

Based on the above findings, we propose an Adaptive Adversarial Logits Pairing training method, which greatly improves the effectiveness of adversarial training. The main contributions are as follows:
\begin{itemize}
  \item We propose a Guided Dropout that tailors features based on feature contributions. It can prevent overfitting while improving the model's robustness.
  \item We propose a method to adaptively adjust the Adversarial Logits Pairing loss.
  \item Our proposed method greatly improves the robustness of the model against adversarial attacks and shows excellent results on multiple datasets.
\end{itemize}

\section{Data Analysis And Motivation}
The Adversarial Logits Pairing is currently the most classic adversarial training method that constrains the activation layer.
The Adversarial Logits Pairing consists of two constraints, among which classification loss is used to constrain the correct classification of the model, and ALP loss constrains the original sample and the adversarial sample to guarantee a similar output.
ALP loss ~\cite{alp} is defined as follows:

\begin{equation}
  ALP\ loss = ||Logits_{clean}-Logits_{adv}||_2^2
\end{equation}

where $Logits_{clean}$ and $Logits_{adv}$ refer to the activation value of the original sample in the Logits layer and the activation value of the adversarial sample in the Logits layer respectively.

We want to explore the mechanism of the ALP method and find a general method for improving adversarial training from the perspectives of the training process and the training target. The data analysis part intends to solve the following two problems:

\begin{itemize}
  \item How does the ALP method affect the model features after training?
  \item Is the training target of the ALP method applicable to all samples?
\end{itemize}


In this section, we define the PGD ~\cite{madry2017towards} as the adversarial attack, and the attack configuration is as follows:
On the MNIST dataset: $\epsilon$ = 0.3, attack step size = 0.01, iteration = 40,
SVHN dataset ~\cite{SVHN}: $\epsilon$ = 12pix, attack step size = 3pix, iteration = 10,
CIFAR dataset ~\cite{cifarkrizhevsky2009learning}: $\epsilon$ = 8pix, attack step size = 2pix, iteration = 7.

\subsection{Feature Analysis}

Although the Adversarial Logits Pairing performs well, its mechanism is still unclear. The original author only simply used ALP as an extension of adversarial training, but did not analyze in-depth how this extension improved the effect of adversarial training. 

We use first-order Taylor expansion to further derive the formula of ALP, and give a more intuitive form, through which we can intuitively see how ALP improves adversarial training. We define $f(\cdot)$ as a function to obtain the activation of the Logits layer of the neural network, so ALP can be rewritten as:
\begin{equation}
  \begin{split}
    &ALP\ loss = (f(x_{adv}) - f(x))^T(f(x_{adv}) - f(x)) \\
    &=(f(x+\Delta x) - f(x))^T(f(x+\Delta x) - f(x)) \\
    &=(f(x) + \frac{\partial f(x)}{\partial x} \Delta x - f(x))^T(f(x) + \frac{\partial f(x)}{\partial x} \Delta x - f(x)) \\
    &=(\frac{\partial f(x)}{\partial x} \Delta x)^T(\frac{\partial f(x)}{\partial x} \Delta x) \\
  \end{split}
\end{equation}

where $\Delta x$ is adversarial perturbations, adversarial perturbations can be approximated as: 
\begin{equation}
  \Delta x = \alpha  \frac{\partial Loss}{\partial x} 
\end{equation}

So, ALP loss can be further expressed as:
\begin{equation}
  \begin{split}
    &ALP\ loss =\alpha^2(\frac{\partial f(x)}{\partial x} \frac{\partial Loss}{\partial x})^T(\frac{\partial f(x)}{\partial x} \frac{\partial Loss}{\partial x}) \\
    &=\alpha^2(\frac{\partial f(x)}{\partial x} \frac{\partial Loss}{\partial f(x)} \frac{\partial f(x)}{\partial x})^T(\frac{\partial f(x)}{\partial x} \frac{\partial Loss}{\partial f(x)} \frac{\partial f(x)}{\partial x}) \\
    &=\alpha^2 \frac{\partial Loss}{\partial f(x)} \frac{\partial f(x)}{\partial x} (\frac{\partial f(x)}{\partial x})^T \frac{\partial f(x)}{\partial x} (\frac{\partial f(x)}{\partial x})^T \frac{\partial Loss}{\partial f(x)}
  \end{split}
  \label{eqa_alp}
\end{equation}

The rewritten ALP loss is mainly composed of $\frac{\partial Loss}{\partial f(x)}$ and $\frac{\partial f(x)}{\partial x}$. Constraining ALP loss means that the product of $\frac{\partial Loss}{\partial f(x)}$ and $\frac{\partial f(x)}{\partial x}$ becomes smaller. The expression of $\frac{\partial Loss}{\partial f(x)}$ is very similar to the Grad-CAM algorithm. Constraint $\frac{\partial Loss}{\partial f(x)}$ can be understood as reducing high-contribution features. $\frac{\partial f(x)}{\partial x}$ can be understood as the gradient of the Logits layer to the input image, and constraint $\frac{\partial f(x)}{\partial x}$ means to make the input gradient smoother.

We want to further quantify the impact of feature contributions on the robustness of the model,
so we define the value of $C$ as the contribution of the feature to the training target,

\begin{equation}
    C = mean(abs(\frac{\partial Loss}{\partial A_{fc}}))
\end{equation}

where $mean(\cdot)$ represents the average function, $abs(\cdot)$ represents the absolute value function, and $A_{fc}$ refers to the activation value of the layer before the Logits layer. 
Because the output of the Logits layer contains class information, we choose the activation value of the previous layer of the Logits layer to calculate the feature contribution.

\begin{figure}[t]
  \centering
  \begin{minipage}[c]{0.08\textwidth}
    SVHN
  \end{minipage}
  \begin{minipage}[c]{0.40\textwidth}
    \includegraphics[width=0.85\linewidth]{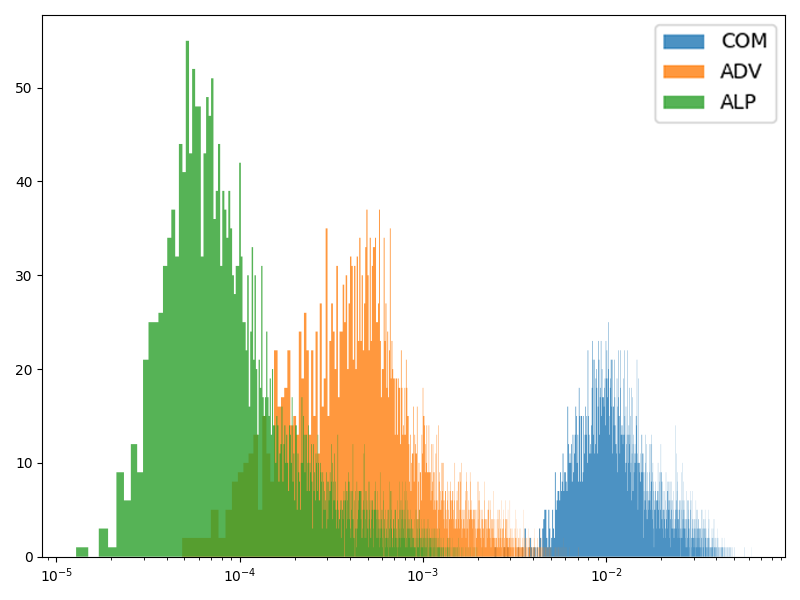}
  \end{minipage}

  \begin{minipage}[c]{0.08\textwidth}
    Cifar10
  \end{minipage}
  \begin{minipage}[c]{0.40\textwidth}
    \includegraphics[width=0.85\linewidth]{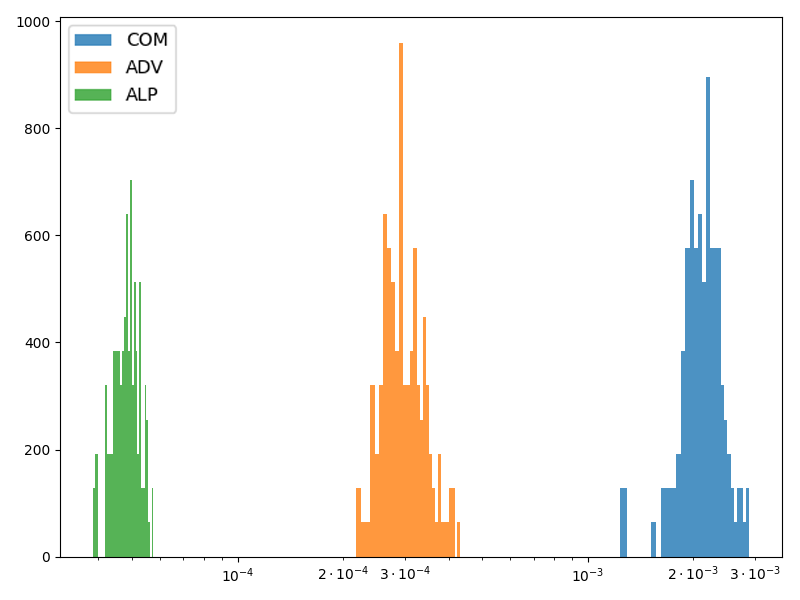}
  \end{minipage}
    \caption{Distribution histogram of the feature contribution of different training algorithms. 
    COM represents general training; ADV represents adversarial training; ALP represents adversarial logits pairing training.
    The horizontal axis represents feature contribution, and the vertical axis represents frequency.}
    \label{feature_bf}
\end{figure}

We calculate the contribution of the previous layer of Logits after convergence in the SVHN dataset and Cifar10 dataset. 
As the robustness of the COM, ADV, and ALP algorithms increases in turn, it can be seen from Fig.~\ref{feature_bf} that as the robustness increases, the model shows fewer high-contribution features

For feature analysis, GradCAM~\cite{selvaraju2017grad} is a commonly used tool, which can visualize the contribution of features. GradCAM provides the activation map with the gradients and activations of the models:

\begin{equation}
  Grad\ CAM = relu(\frac{\partial Loss_{c}}{\partial A_{i}}*A_{i})
\end{equation}
where $relu(\cdot)$ represents the relu function, $Loss_{c}$ represents the classification loss of the model for category $c$, and $A_{i}$ refers to the activation value of the layer i.

\begin{figure}[t]
  \centering
  \includegraphics[width=8cm]{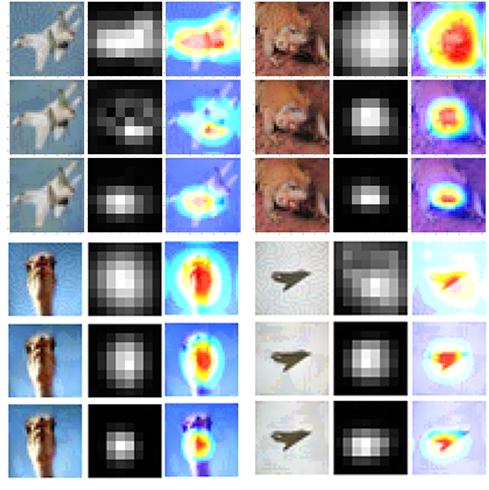}
  \caption{Changes in the activation maps of adversarial samples in three different training methods. (From top to bottom: activation maps of general training, adversarial training and ALP method)}
  \label{cam_bf}
\end{figure}

We visualize the adversarial sample's activation maps of the three methods: general training, adversarial training and ALP training in Fig.~\ref{cam_bf}.
It can be seen from the figure that as the robustness of the model increases, the areas strongly related to the model gradually decrease.

The difference in the activation maps and the distributions can prove:
\begin{itemize}
  \item Robust model features have the distribution characteristics of few high-contribution.
  \item Strongly related areas during training are also areas of greater concern for neural networks.
\end{itemize}

The changes in the activation maps and the distributions of contribution show a high degree of consistency. The fewer high-contribution features cause the model to focus on the most critical areas in the image. Thereby the model reducing the impact of other areas on the model output after being attacked by adversarial samples.

This phenomenon inspires us to make the model focus more on the most important part of the image during the training process so as to improve the robustness of the model.

\subsection{Sample Optimization Analysis}
In the previous section, we find that the robust model presents the distribution characteristics of fewer high-contribution features. The highs contribution feature will bring strong confidence, and ALP is a typical multi-loss collaborative method, which is prone to cause interactions among losses. So, we want to know whether the fewer high-contribution feature distribution will affect the confidence of the model.

\begin{figure}[t]
  \centering
  \begin{minipage}[c]{0.05\textwidth}
  MNIST  
  \end{minipage}
  \begin{minipage}[c]{0.2\textwidth}
  \centering
  \includegraphics[width=3.5cm]{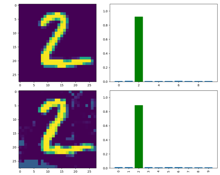}
  \end{minipage}
  \begin{minipage}[c]{0.2\textwidth}
  \centering
  \includegraphics[width=3.5cm]{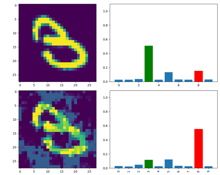}
  \end{minipage}

  \centering
  \begin{minipage}[c]{0.05\textwidth}
  SVHN  
  \end{minipage}
  \begin{minipage}[c]{0.2\textwidth}
  \centering
  \includegraphics[width=3.5cm]{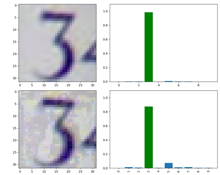}
  \end{minipage}
  \begin{minipage}[c]{0.2\textwidth}
  \centering
  \includegraphics[width=3.5cm]{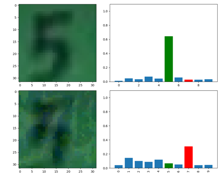}
  \end{minipage}

  \centering
  \begin{minipage}[c]{0.05\textwidth}
  Cifar10
  \end{minipage}
  \begin{minipage}[c]{0.2\textwidth}
  \centering
  \includegraphics[width=3.5cm]{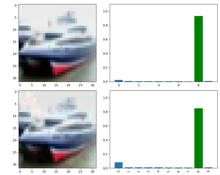}
  \end{minipage}
  \begin{minipage}[c]{0.2\textwidth}
  \centering
  \includegraphics[width=3.5cm]{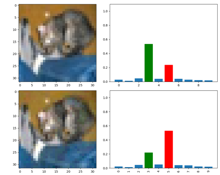}
  \end{minipage}
  \caption{On the left is a class of data with correct classification of the original and adversarial samples, on the right is a class of data with correct classification of the original samples but wrong classification of adversarial samples, the first row of each group of figures is the original sample, the second row is the adversarial sample, green bar is the confidence score of the ground truth, and the red bar is the confidence score of the attack target class.}
  \label{fig_2_example}
\end{figure}

We choose the ALP model and find that the samples mainly present two states in the three data sets of MNIST, SVHN and Cifar10.
One type of successfully defended samples, 
proves the ALP model’s validity in the defense of the adversarial sample with equal high confidence as with the original samples, 
while the other type of failed defended samples is generally accompanied by lower confidence. The two types of samples are shown in Fig.~\ref{fig_2_example}.


As for the original intention of adversarial defense and ALP, we want to avoid the phenomenon, which the original sample is correctly classified and the adversarial sample is incorrectly classified. We define a type of sample 
with correct classification of the original sample but wrong classification of the adversarial sample as the Inconsistent Set. 
Similarly, all correctly classified samples are defined as Consistent Set. For Inconsistent Set samples obviously violating the ALP training target, we want to explore why such samples fail to be defended by the ALP model.


Because ALP focuses more on the constraints of confidence score, we decide to observe the confidence score in the Inconsistent Set and use the samples in the Consistent Set to compare in order to explore the characteristics of the samples that failed to defend. We separately count the clean sample's ground-truth confidence, adversarial sample's ground-truth confidence, classification loss, and ALP loss of the two types of samples in the test set, as shown in the following Table.~\ref{table_2_example}.

\begin{table}[t]
  \begin{center}
  \resizebox{0.5\textwidth}{6mm}{
  \begin{tabular}{|c|c|c|c|c|}
  \hline
  MNIST & $Avg Probs_{clean}$ & $Avg Probs_{adv}$ & Classification loss & ALP loss \\
  \hline
  Consistent Set & \textbf{94.61\%} & \textbf{92.58\%} & 0.3262 & 0.1443 \\ 
  \hline
  Inconsistent Set & 60.94\% & 21.87\% & \textbf{4.1833} & \textbf{2.2442} \\
  \hline
  \end{tabular}}
  \end{center}

  \begin{center}
  \resizebox{0.5\textwidth}{6mm}{
  \begin{tabular}{|c|c|c|c|c|}
  \hline
  SVHN & $Avg Probs_{clean}$ & $Avg Probs_{adv}$ & Classification loss & ALP loss \\
  \hline
  Consistent Set & \textbf{86.25\%} & \textbf{52.37\%} & 5.1866 & 0.9199 \\ 
  \hline
  Inconsistent Set & 54.61\% & 11.22\% & \textbf{7.0301} & \textbf{3.0740} \\
  \hline
  \end{tabular}}
  \end{center}

  \begin{center}
  \resizebox{0.5\textwidth}{6mm}{
  \begin{tabular}{|c|c|c|c|c|}
  \hline
  Cifar10 & $Avg Probs_{clean}$ & $Avg Probs_{adv}$ & Classification loss & ALP loss \\
  \hline
  Consistent Set & \textbf{89.53\%} & \textbf{69.81\%} & 2.0889 & 0.5236 \\ 
  \hline
  Inconsistent Set & 59.88\% & 14.34\% & \textbf{6.2318} & \textbf{2.7438} \\
  \hline
  \end{tabular}}
  \end{center}
  \caption{Confidence score, classification loss, and alp loss before and after the adversarial attack in Inconsistent Set and Consistent Set.}
  \label{table_2_example}
\end{table}


The above data analysis shows that the following differences exist between Inconsistent Set and Consistent Set:
\begin{itemize}
  \item Consistent Set's clean sample average confidence and adversarial sample average confidence are much higher than that of Inconsistent Set.
  \item Inconsistent Set's classification loss and ALP loss are much higher than that of Consistent Set.
  \item The percentage of ALP loss taking over total loss in Consistent Set is less than that in Inconsistent Set.
\end{itemize}



In the data analysis of Consistent Set and Inconsistent Set, we reckon that Inconsistent Set has an adverse effect on the training target of adversarial training, so we decide to observe the differences between Inconsistent Set and Consistent Set during the entire ALP training process.

\begin{figure}[h]
  \begin{center}
     \includegraphics[width=0.85\linewidth]{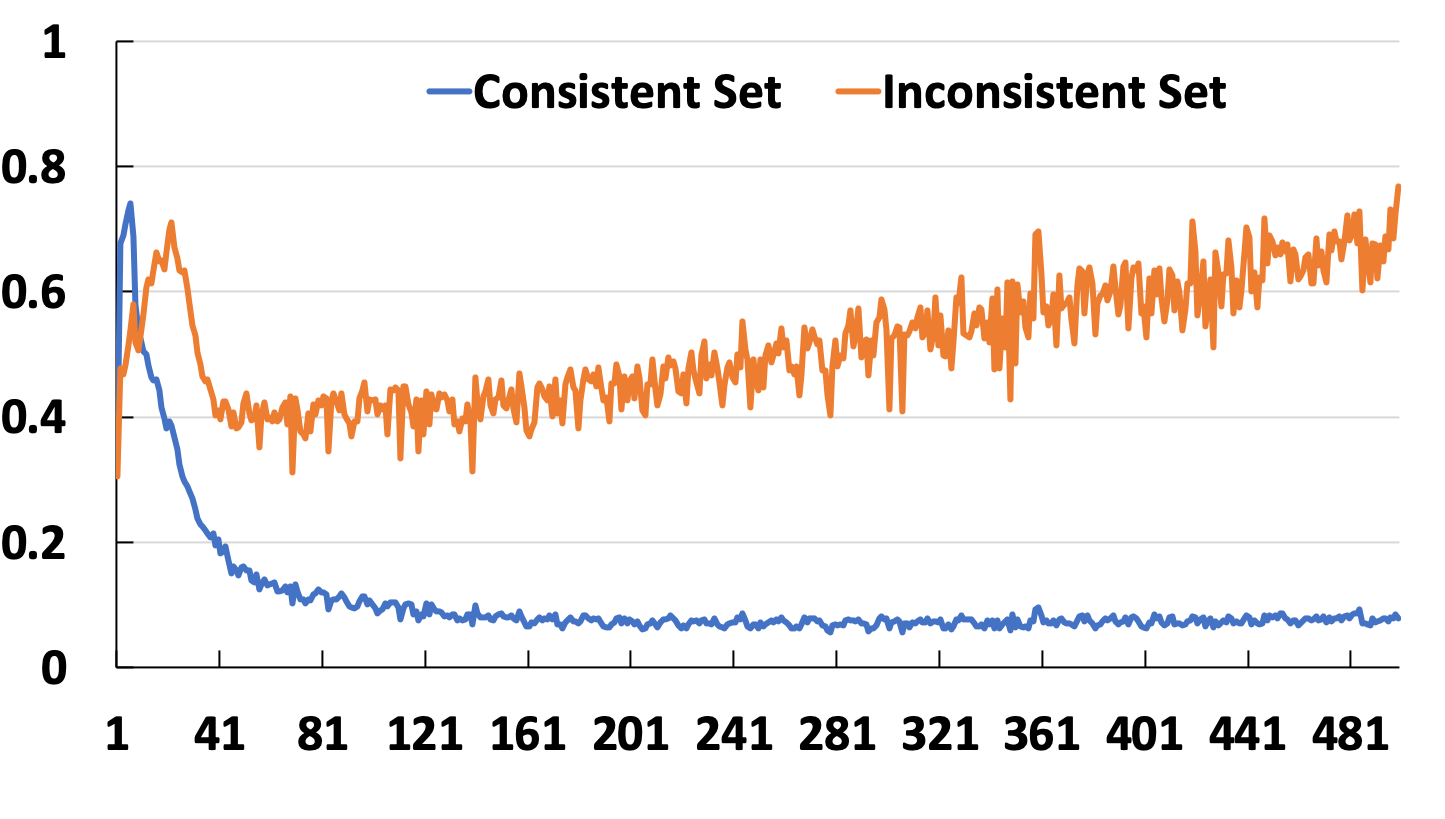}
  \end{center}
    \caption{ALP loss value changes of Inconsistent Set and Consistent Set with the increase of training times.}
    \label{Incon_and_consis}
\end{figure}

We record the ALP loss value of the model for each sample of the test set after the end of each epoch in the training process and divide the samples into Consistent Set and Inconsistent Set according to the situation of the model convergence. Then we print the average ALP loss of each epoch of Consistent Set and Inconsistent Set, as shown in Fig.~\ref{Incon_and_consis}.

It can be seen from the figure that during an ALP training process, the ALP loss of Consistent Set gradually decreases, and the robustness gradually increases under the constraint of ALP loss. The ALP loss of Inconsistent Set is gradually increasing. Obviously, the constraint of ALP loss is too strong for Inconsistent Set, which affects the convergence of such samples.

Therefore, although the ALP algorithm plays a certain role in defense, it does not have a good enough training target, which hampers the overall defense effect of the model. This phenomenon inspired us to design an ALP loss that can treat Consistent Set and Inconsistent Set differently so as to achieve a better training effect.

\section{Method}

\begin{figure*}[t]
  \begin{center}
     \includegraphics[width=0.95\linewidth]{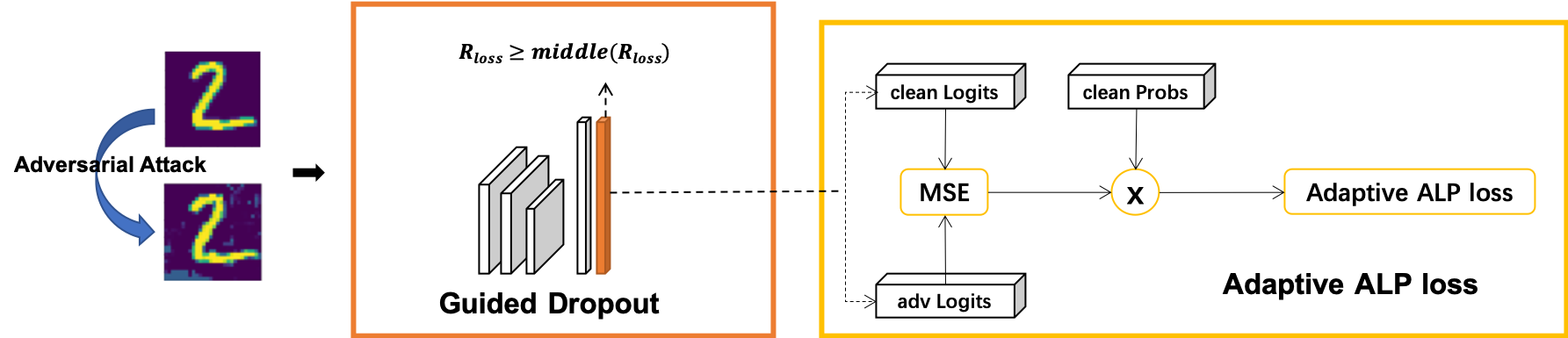}
  \end{center}
      \caption{AALP training framework. 
      The upper is original sample and the lower is the adversarial sample.
      Guided Dropout is in the layer before the Logits layer.}
\end{figure*}

In this section, we will solve the problems in the ALP method from two perspectives:
(1) For optimizing the training process, we want to give the model a higher priority to train high-contribution features. (2) In order to alleviate the negative impact of the ALP loss on Inconsistent Samples at the training target, we want to design an adaptive loss that can coordinate different samples.

\subsection{Adaptive Feature Optimization}
According to the data analysis, reducing high-contribution features will help the model improve robustness. This phenomenon motivates us to design an algorithm that is intended to train high-contribution features.


In terms of model structure, the dropout layer is the most suitable model structure to achieve this demand. The initial design of dropout is to solve the over-fitting phenomenon of the model ~\cite{dropout}. By randomly cutting the model features, the quality of the model features and the generalization of the model are improved. In recent years, many algorithms have improved the performance of the model by controlling the cutting process of the dropout layer, such as combining attention with dropout ~\cite{att_dropout}. Therefore, we want to combine dropout and feature contribution to make the dropout layer conform to certain rules for feature cutting, so that the models have fewer high-contribution features, thereby improving the robustness of the model.

We use the characteristics of the dropout layer to selectively tailor the neural network. By keeping the highly contributed features and cutting the weakly contributed features, we force the model to focus on the highly contributed features during the training process. Specifically, in forwarding propagation, we first calculate the contribution scores between each neuron in the previous layer of the Logits layer, and then retain the first 50\% of neurons and discard the rest 50\% of neurons. Then, we update the parameters of the model after trimming to realize Guided Dropout.

We combine the expansion of ALP loss in the Equ.~\ref{eqa_alp} and the GradCAM algorithm to define $R_{loss}$ to measure feature contribution.

\begin{equation}
  R_{loss} = mean(abs(\frac{\partial Loss}{\partial A_{fc}}*A_{fc}))
\end{equation}

where $A_{fc}$ represents the activation value of the previous layer of Logits, $abs(\cdot)$ represents the absolute value function, and $mean(\cdot)$ represents the average function. The resulting $R_{loss}$ is a vector with the same shape as $A_{fc}$, which stores the contribution scores of each feature of the $A_{fc}$ layer with the current round of data.

\begin{equation}
  Guided\ Dropout = R_{loss} \ge middle(R_{loss})
  \label{GD}
\end{equation}

where $middle(\cdot)$ represents the median function. We want to keep the high-contribution features in the training process to continue training so that the contribution of those features can be reduced. 
After making comparison in Equ.~\ref{GD}, we define that the part of $R_{loss}$ greater than the median is 1, and the part less than the median is 0.
Then the vector is regarded as the mask of the dropout, so as to realize the Guided Dropout.

We analyze the reason why the Guided Dropout module works. We believe that high-contribution features will have higher $l_2\ loss$ due to their larger weight in the fully connected layer. When they are trained separately, $l_2\ loss$ will restrict their weight to decrease, thus achieving our goal of reducing the number of high-contribution features.

\subsection{Adaptive Sample Weighting}

As analyzed above, the ALP loss does not have a good effect on all samples. In the process of the ALP loss restraining the Inconsistent Samples, 
we find the normal convergence is sometimes affected.
So, we propose a training method called Adaptive ALP loss to adaptively control the ALP loss weight of different samples in the training process. 
According to the experimental results in the previous section, it is noted that the Inconsistent Set has a large impact and the confidence of the Inconsistent Set is low. In order to weaken the negative impact of ALP loss, we propose to use the confidence score to weight ALP loss.
We first calculate the confidence scores of the original samples and set these confidence scores as the ALP loss weight of the samples in the training round. Adaptive ALP loss can effectively distinguish the Consistent Samples and the Inconsistent Samples, reduce the ALP loss weight of the Inconsistent Samples, and better help against training convergence. At the same time, adaptive ALP loss will overall reduce the negative impact of ALP loss during the entire training process, allowing the model to converge faster.

\begin{equation}
  Adaptive\ ALP\ loss = Probs_{clean}*ALP\ loss
\end{equation}
where $Probs_{clean}$ denotes the clean sample's confidence score of the ground-truth.
The overall loss is calculated as follows:
\begin{equation}
  \begin{aligned}
    Total\ loss = Classification\ Loss + \alpha * Adaptive\ ALP\ loss
  \end{aligned}
\end{equation}
where $\alpha$ is the weight of the Adaptive ALP loss.

\section{Experiment}
In the experimental part, we want to verify whether we have reached the designed goals of the algorithm: 
whether AALP loss treats the Consistent and Inconsistent Sets differently,
and whether the Guided Dropout reduces the contribution of the high-contribution features. 
At the same time, we show the adversarial defense effect and analyze the parameter selection.

\begin{table*}[!th]
  \begin{center}
    \begin{tabular}{|c|c|c|c|c|c|c|c|c|c|c|c|}
      \hline
      DataSet & Method & Clean & FGSM & PGD10 & PGD20 & PGD40 & PGD100 & CW10 & CW20 & CW40 & CW100 \\ \hline
      \multirow{7}*{MNIST}
      & RAW & \textbf{99.18\%} & 7.72\% & 0.49\% & 0.54\% & 0.64\% & 0.53\% & 0.0\% & 0.0\% & 0.0\% & 0.0\% \\ \cline{2-12}
      & Madry adv &	99.00\% & 96.92\%	& 95.69\% & 95.52\% & 94.14\% & 92.49\% & 96.28\% & 95.85\% & 94.53\% & 93.78\% \\ \cline{2-12}
      & Kannan ALP & 98.77\% & 97.68\% & 96.66\% & 96.58\% & 96.37\% & 94.50\% & 96.96\% & 97.07\% & 96.73\% & 96.16\% \\ \cline{2-12}
      & TRADES & 98.48\% & 98.34\% & 96.90\% & 96.68\% & 96.63\% & 95.35\% & 97.09\% & 97.09\% & 96.96\% & 95.76\% \\ \cline{2-12}
      & $AALP_{gd}$ & 98.16\% & 98.17\% & 97.13\% & 97.04\% & 96.71\% & 95.72\% & 96.92\% & 97.08\% & 96.78\% & 96.39\% \\ \cline{2-12}
      & $AALP_{sw}$ & 98.16\% & 97.77\% & 97.05\% & 96.82\% & 96.66\% & 95.87\% & 97.18\% & 97.15\% & 96.87\% & 96.76\% \\ \cline{2-12}
      & AALP & 98.49\% & \textbf{98.19\%} & \textbf{97.22\%} & \textbf{97.13\%} & \textbf{97.15\%} & \textbf{96.32\%} & \textbf{97.28\%} & \textbf{97.54\%} & \textbf{97.26\%} & \textbf{96.89\%} \\ \hline \hline
      \multirow{7}*{SVHN}
      & RAW & \textbf{95.57\%} & 8.85\% & 2.23\% & 2.22\% & 2.25\% & 2.17\% & 0.83\% & 0.76\% & 0.71\% & 0.66\% \\ \cline{2-12}
      & Madry adv & 86.41\% & 48.53\% & 31.51\% & 27.44\% & 25.75\% & 25.02\% & 38.41\% & 35.32\% & 33.93\% & 33.04\% \\ \cline{2-12}
      & Kannan ALP & 85.28\% & 53.47\% & 40.05\% & 36.55\% & 35.07\% & 34.15\% & 44.82\% & 42.48\% & 41.52\% & 40.86\% \\ \cline{2-12}
      & TRADES & 84.05\% & 61.14\% & 42.71\% & 40.42\% & 39.46\% & 38.78\% & 46.77\% & 44.83\% & 44.00\% & 43.54\% \\ \cline{2-12}
      & $AALP_{gd}$ & 86.23\% & 59.18\% & 44.05\% & 42.57\% & 42.27\% & 41.23\% & 46.47\% & 44.96\% & 44.51\% & 44.05\% \\ \cline{2-12}
      & $AALP_{sw}$ & 84.69\% & \textbf{62.83\%} & 44.30\% & 40.99\% & 40.99\% & 39.94\% & 47.78\% & 45.87\% & 44.68\% & 43.80\% \\ \cline{2-12}
      & AALP & 84.13\% & 57.67\% & \textbf{45.05\%} & \textbf{43.35\%} & \textbf{42.60\%} & \textbf{42.18\%} & \textbf{47.92\%} & \textbf{47.08\%} & \textbf{46.77\%} & \textbf{46.57\%} \\ \hline \hline
      \multirow{7}*{Cifar10}
      & RAW & \textbf{90.51\%} & 14.24\% & 5.93\% & 5.80\% & 5.97\% & 6.04\% & 1.92\% & 1.88\% & 1.73\% & 1.53\% \\ \cline{2-12}
      & Madry adv & 83.36\% & 63.28\%	&  49.29\% & 45.25\% & 44.85\% & 44.72\% & 46.20\% & 45.16\% & 44.76\% & 44.65\% \\ \cline{2-12}
      & Kannan ALP & 82.80\% & 64.96\% &	52.79\% & 49.04\% & 48.81\% & 48.60\% & 52.81\% & 52.13\% & 51.79\% & 51.65\% \\ \cline{2-12}
      & TRADES & 82.73\% & 64.17\% & 53.47\% & 52.52\% & 52.38\% & 52.29\% & 59.74\% & 58.91\% & 58.71\% & 58.66\% \\ \cline{2-12}
      & $AALP_{gd}$ & 80.50\% & 65.00\% & 54.78\% & 51.69\% & 51.51\% & 51.40\% & 59.00\% & 58.45\% & 58.31\% & 58.23\% \\ \cline{2-12}
      & $AALP_{sw}$ & 82.37\% & \textbf{65.69\%} & 53.96\% & 50.25\% & 49.91\% & 49.74\% & 53.35\% & 52.51\% & 52.33\% & 52.12\% \\ \cline{2-12}
      & AALP & 80.45\% & 65.42\% & \textbf{55.23\%} & \textbf{52.59\%} & \textbf{52.41\%} & \textbf{52.38\%} & \textbf{60.00\%} & \textbf{59.47\%} & \textbf{59.34\%} & \textbf{59.30\%} \\ \hline 

    \end{tabular}
    \caption{Comparison of defense performance with Madry, Kannan and others under different adversarial attack methods on muti datasets.}
    \label{table_mnist} 
    \label{table_svhn}
    \label{table_cifar} 
  \end{center}

\end{table*}

\subsection{Experimental Setting}

The Adaptive ALP experiment was carried out on three datasets and models, \emph{i.e.} LeNet ~\cite{lenet} trained based on MNIST dataset, ResNet9 ~\cite{resnet} trained based on SVHN dataset and ResNet32 trained based on Cifar10 dataset.
MNIST basic configuration: Backbone network is LeNet. We use Adam optimization ~\cite{adam} and make lr = 0.0001. PGD40 is used to create the adversarial samples and attack step size is 0.01, $\epsilon$ = 0.3.
SVHN basic configuration: backbone network is ResNet9. We use Adam optimization and make lr = 0.0001. We select PGD10 to generate the adversarial samples and attack step size is 3pix, $\epsilon$ = 12pix.
The basic configuration of Cifar10: Backbone network is ResNet32. We use Adam optimization and make lr = 0.00001. We select PGD7 to generate the adversarial samples and attack step size is 2pix, $\epsilon$ = 8pix.

\subsection{Performance on Adversarial Defense}


We conduct a defense effect test on three datasets, comparing Madry's adversarial training algorithm, Kannan's ALP algorithm and Zhang's TRADES algorithm.
We choose FGSM, PGDX, CWX (X represents the number of iterations) as the comparative attack algorithms.





On the MNIST dataset in Table.~\ref{table_mnist}, AALP performs well. In the case of very limited improvement space, the AALP algorithm still improves the defense effect by nearly 1\% compared to the ALP algorithm in multiple attack modes.
AALP improves by nearly 2\% compared to the ALP algorithm, especially under the attack of PGD100.

On the SVHN dataset in Table.~\ref{table_svhn}, we can see that AALP has significantly improved the defense of the two algorithms: PGD and CW. In the defense of the FGSM algorithm, the AALP algorithm performs slightly inferior to the TRADES algorithm, but $AALP_{sw} $ can still maintain the same or even slightly higher results as the TRADES algorithm.

On the Cifar10 dataset in Table.~\ref{table_cifar}, the performance of the AALP algorithm is similar to that of the SVHN dataset, and the overall defense of the iterative attack algorithm has been significantly improved. For the single step attack, FGSM improves slightly, but it is not significant.


Compared with the SOTA method TRADES, we can obviously observe that the AALP training method has significantly improved the adversarial defense effect, especially in the iterative attack.

Compared with ALP, AALP has made considerable progress. Since the ALP algorithm performs well on large data sets and complex models, the AALP algorithm with improved Guided Dropout and Adaptive Sample Weighting will definitely perform better on large data sets. 
In this article, because the major part is dedicated to the explanation of the internal reasons for the
AALP algorithm improvement, the AALP algorithm’s performance on large data sets is not covered.

\subsection{Parameter Analysis}




The algorithm we propose has one main parameter that can be adjusted, namely $\alpha$, which controls the loss ratio.
We selected the SVHN dataset for parameter analysis.

\begin{figure}[t]
  \centering
  \includegraphics[width=5.5cm]{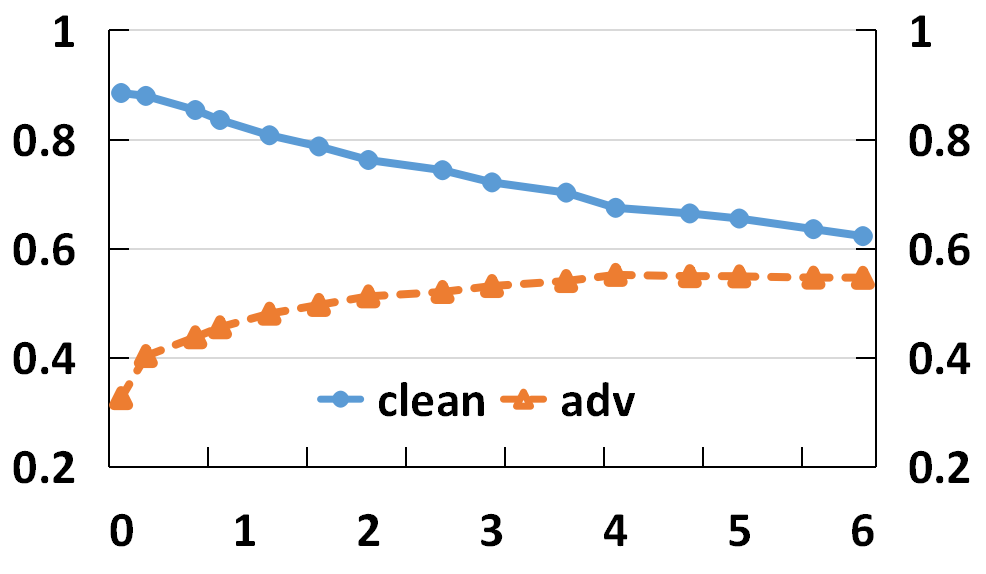}
  \caption{AALP's defense performance under different weight parameter configurations, the figure shows the changes of $\alpha$}
  \label{para}
\end{figure}

In Fig.~\ref{para}, the change of $\alpha$ has obvious effects on defense. When we increase $\alpha$, it will obviously improve the defense effect, but it will also lose the clean classification accuracy, 
so we choose the final result $\alpha = 1.0$, which is better for the overall result.

\subsection{Fewer High-Contribution Features}

In the previous data analysis, we found that models with stronger robustness usually have fewer high-contribution features, and these high-contribution features are usually concentrated in the parts that are crucial for classification. 

\begin{figure}[t]
  \centering
  \begin{minipage}[c]{0.08\textwidth}
    SVHN
  \end{minipage}
  \begin{minipage}[c]{0.4\textwidth}
    \includegraphics[width=0.9\linewidth]{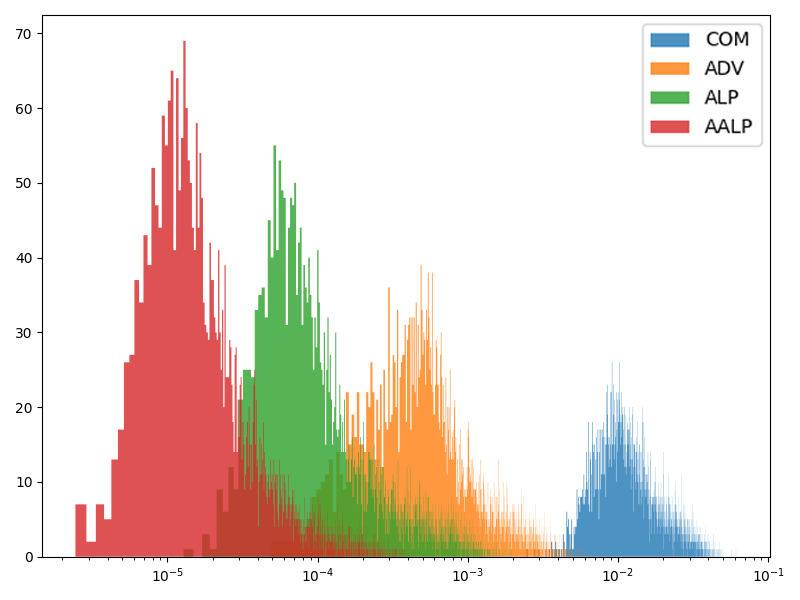}
  \end{minipage}

  \begin{minipage}[c]{0.08\textwidth}
    Cifar10
  \end{minipage}
  \begin{minipage}[c]{0.4\textwidth}
    \includegraphics[width=0.9\linewidth]{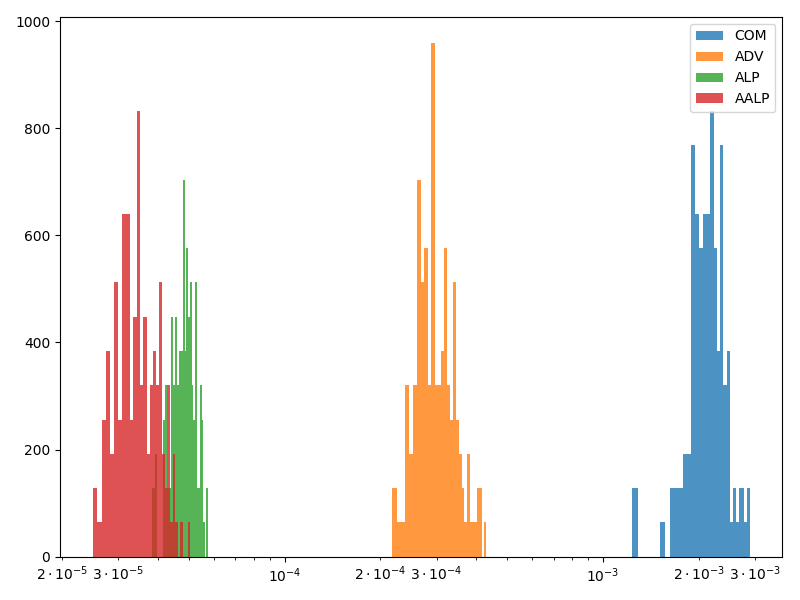}
  \end{minipage}
    \caption{Distribution histogram of the feature contribution of different training algorithms.
    COM represents general training; ADV represents adversarial training; ALP represents adversarial logits pairing training; 
    AALP represents adaptive adversarial logits pairing training. 
    The horizontal axis represents feature contribution, and the vertical axis represents frequency}
    \label{feature_af}
\end{figure}

In Fig.~\ref{feature_af}, we also add the feature contribution of the AALP model to the comparison and find that the model trained by the AALP algorithm has more characteristics of fewer high-contribution features.

\begin{figure}[t]
  \begin{center}
    \includegraphics[width=0.85\linewidth]{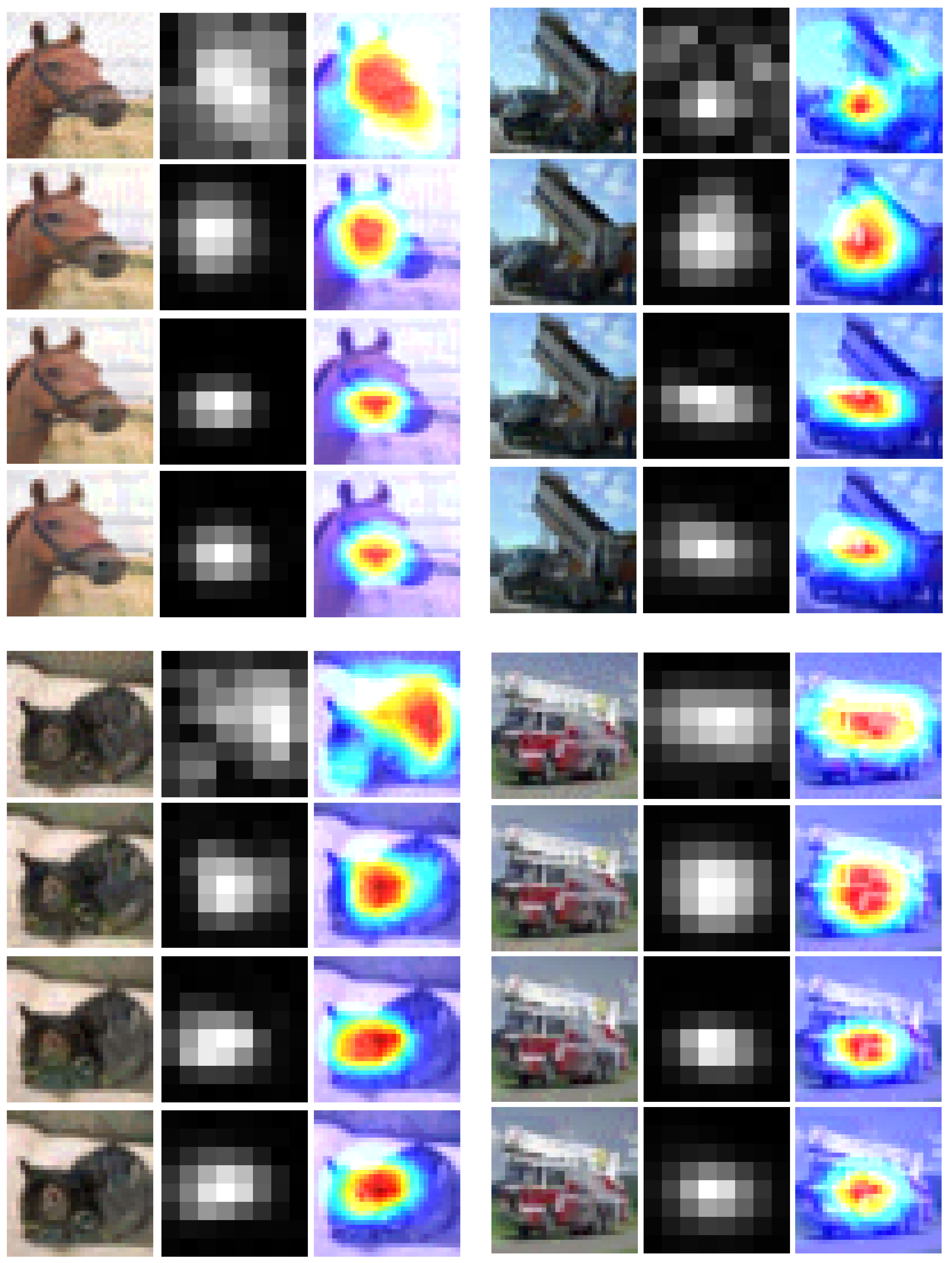}
 \end{center}
  \caption{Changes of activation maps of adversarial samples in the different training methods. (From top to bottom: activation maps for general training, adversarial training, ALP method and AALP method)}
  \label{cam_af}
\end{figure}

In Fig.~\ref{cam_af}, we compare the three defense algorithms ADV, ALP and AALP and their differences in the activation map, finding that the model trained by the AALP algorithm is more willing to focus on the parts that are important for classification.

\begin{figure}[h]
  \begin{center}
     \includegraphics[width=0.9\linewidth]{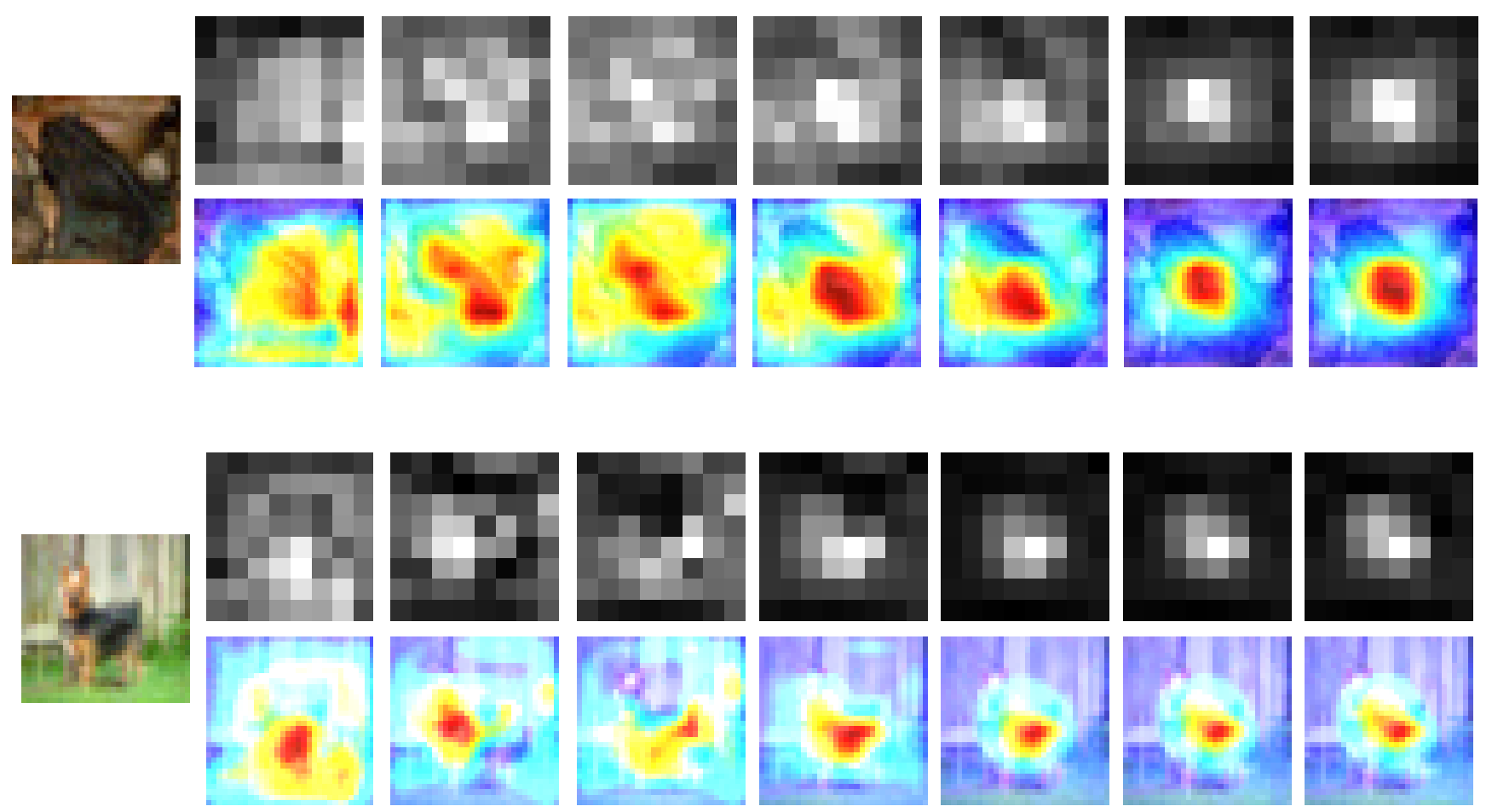}
  \end{center}
    \caption{Changes of activation maps in the process of training AALP.}
    \label{cam_dur}
  \end{figure}

At the same time, we also observe changes in the distribution of features throughout the AALP training process. We train an AALP model and save a model every 10000 steps. We record the feature distribution of each storage point and the activation map of the model under the storage point to obtain Fig.~\ref{feature_dur} and Fig.~\ref{cam_dur}:

\begin{figure}[h]
  \begin{center}
     \includegraphics[width=0.85\linewidth]{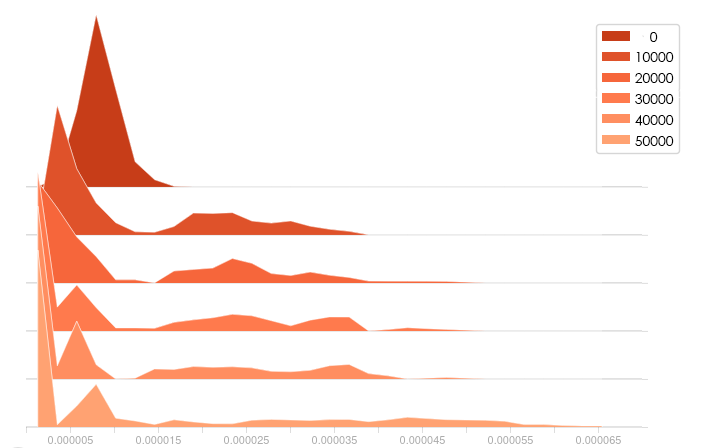}
  \end{center}
    \caption{Changes of feature distributions in the process of training AALP.}
    \label{feature_dur}
  \end{figure}

It can be seen that as the training progresses, the distribution of model features gradually shows a less and less high-contribution trends, and the activation map is also increasingly focused on the key information of the picture.

We also want to test whether Guided Dropout can play a role in other training methods. We use the Resnet9 to train on the SVHN dataset, add the Guided Dropout module to general training and adversarial training(ADV), and observe its working effect.

\begin{table}[t]
  \begin{center}
  \begin{tabular}{|c|c|c|c|}
  \hline
   & clean & PGD10 & CW10 \\
  \hline
  Raw	& \textbf{95.57\%} & 2.23\% & 0.0\% \\ 
  \hline
  Raw+gd & 94.36\% & \textbf{3.17\%} & \textbf{0.04\%} \\
  \hline
  \hline
  ADV & 86.41\% & 31.51\% & 24.03\% \\
  \hline
  ADV+gd & \textbf{95.07\%} & \textbf{35.34\%} & \textbf{34.51\%} \\
  \hline
  \hline
  ALP & 85.28\% & 40.05\% & 31.17\% \\
  \hline
  $AALP_{gd}$ & \textbf{86.23\%} & \textbf{44.05\%} & \textbf{37.57\%} \\
  \hline
  \end{tabular}
  \end{center}
  \caption{Cooperation between Guided Dropout and other algorithms.}
\end{table}


From the data, it can be conducted that the addition of Guided Dropout has played a huge role in defending the adversarial samples. After adding the Guided Dropout module, the ADV model and ALP model have significantly improved the defense of PGD and CW algorithms. The ADV model has also significantly improved the classification of clean samples.
However, Guided Dropout does not seem to work for general training. It makes the accuracy of clean samples slightly lower. Although the defense adversarial samples have improved, they are of little availability.

\subsection{The Changes of Consistent Set and Inconsistent Set}

\begin{figure}[t]
  \centering
  \begin{minipage}[c]{0.2\textwidth}
  \end{minipage}
  \begin{minipage}[c]{0.2\textwidth}
  \centering
  ALP
  \end{minipage}
  \begin{minipage}[c]{0.2\textwidth}
  \centering
  AALP
  \end{minipage}

  \centering
  \begin{minipage}[c]{0.05\textwidth}
  MNIST  
  \end{minipage}
  \begin{minipage}[c]{0.2\textwidth}
  \centering
  \includegraphics[width=3.5cm]{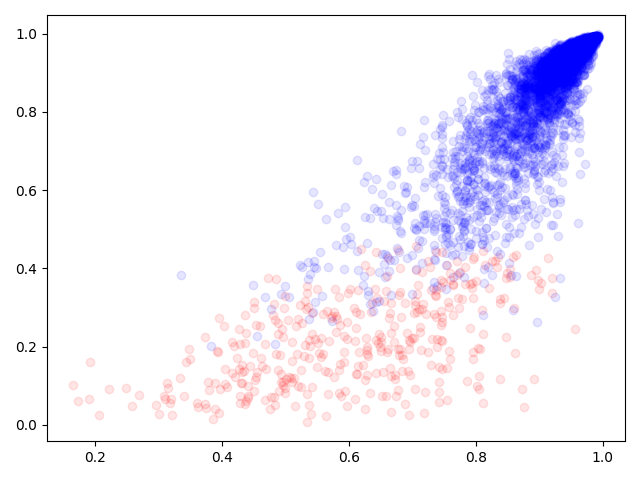}
  \end{minipage}
  \begin{minipage}[c]{0.2\textwidth}
  \centering
  \includegraphics[width=3.5cm]{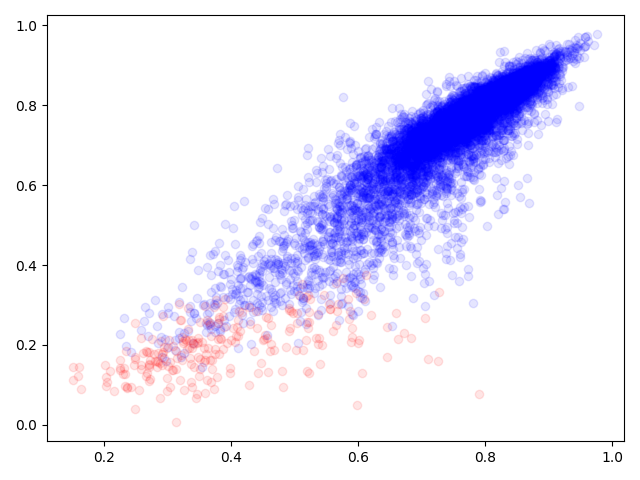}
  \end{minipage}

  \centering
  \begin{minipage}[c]{0.05\textwidth}
  SVHN
  \end{minipage}
  \begin{minipage}[c]{0.2\textwidth}
  \centering
  \includegraphics[width=3.5cm]{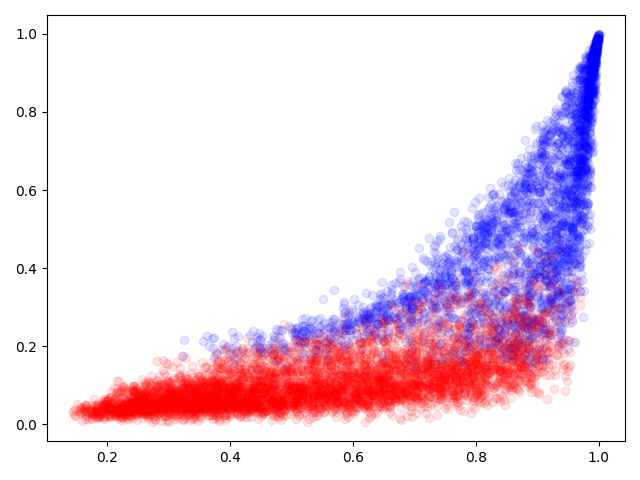}
  \end{minipage}
  \begin{minipage}[c]{0.2\textwidth}
  \centering
  \includegraphics[width=3.5cm]{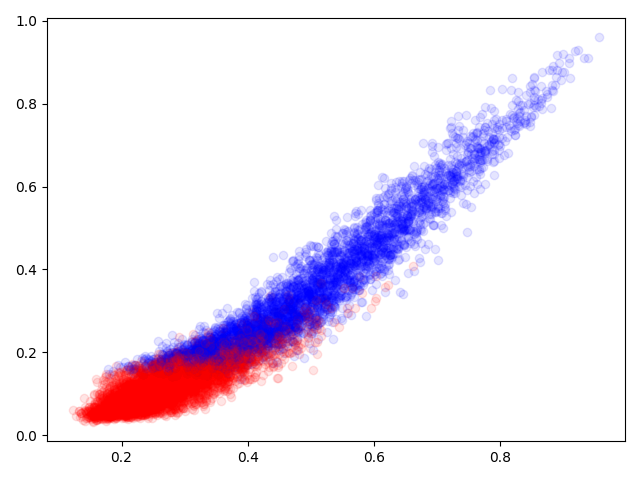}
  \end{minipage}

  \centering
  \begin{minipage}[c]{0.05\textwidth}
  Cifar10
  \end{minipage}
  \begin{minipage}[c]{0.2\textwidth}
  \centering
  \includegraphics[width=3.5cm]{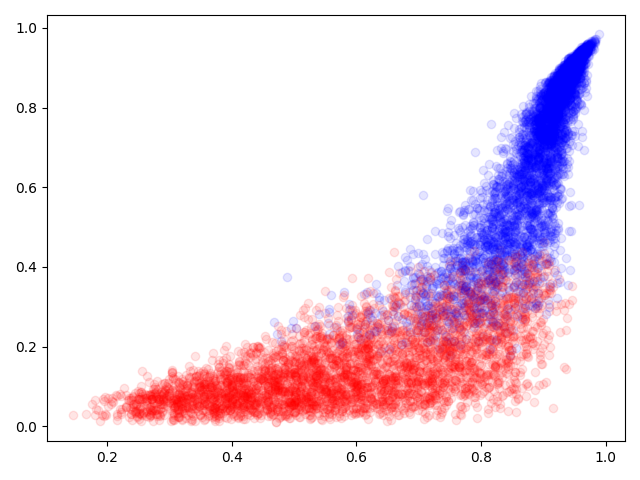}
  \end{minipage}
  \begin{minipage}[c]{0.2\textwidth}
  \centering
  \includegraphics[width=3.5cm]{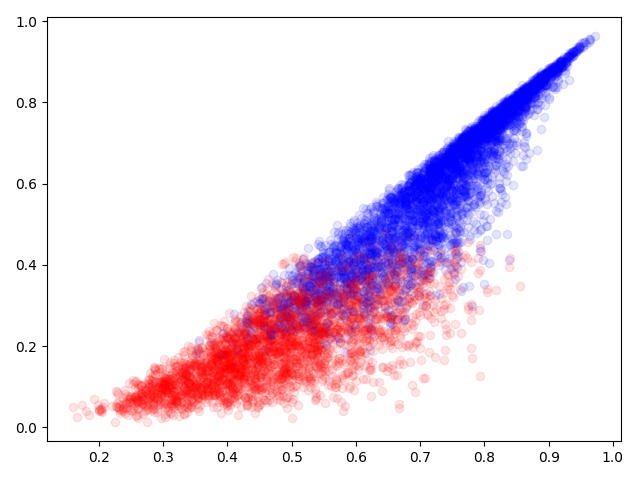}
  \end{minipage}
  \caption{The confidence distribution map of clean samples and adversarial samples in the test set, the blue sample points are Consistent Samples, and the red sample points are Inconsistent Samples.}
  \label{points}
\end{figure}

We compare the changes in the Consistent Set and the Inconsistent Set of the model and the changes in the number of samples of these two types after using the AALP method.
We set the clean sample's confidence scores to the x-axis and the adversarial sample's confidence scores to the y-axis to show the changes of the samples in the entire test set.
In Fig.~\ref{points}, the blue sample points are Consistent Samples, and the red sample points are Inconsistent Samples.

Through the comparison of the three datasets, it can be found that the sample points of the model trained by AALP tend to be distributed on the straight line $y=x$.
It shows that adversarial attack has a lower impact on the output of the AALP model's confidence, and the AALP model has stronger robustness.

Then we compare the changes in the proportion of the Inconsistent Samples in the three datasets.

\begin{table}[t]
  \begin{center}
  \begin{tabular}{|c|c|c|}
  \hline
   & ALP Consistent Samples & AALP Consistent Samples \\
  \hline
  MNIST	& 94.91\% & \textbf{96.03\%} \\ 
  \hline
  SVHN & 30.89\% & \textbf{36.58\%} \\
  \hline
  Cifar10 & 41.61\% & \textbf{43.78\%} \\
  \hline
  \end{tabular}
  \end{center}
  \caption{Consistent Set samples proportion changes after using the AALP method.}
  \label{proportion}
\end{table}





\begin{figure}[!t]
  \begin{center}
     \includegraphics[width=0.8\linewidth]{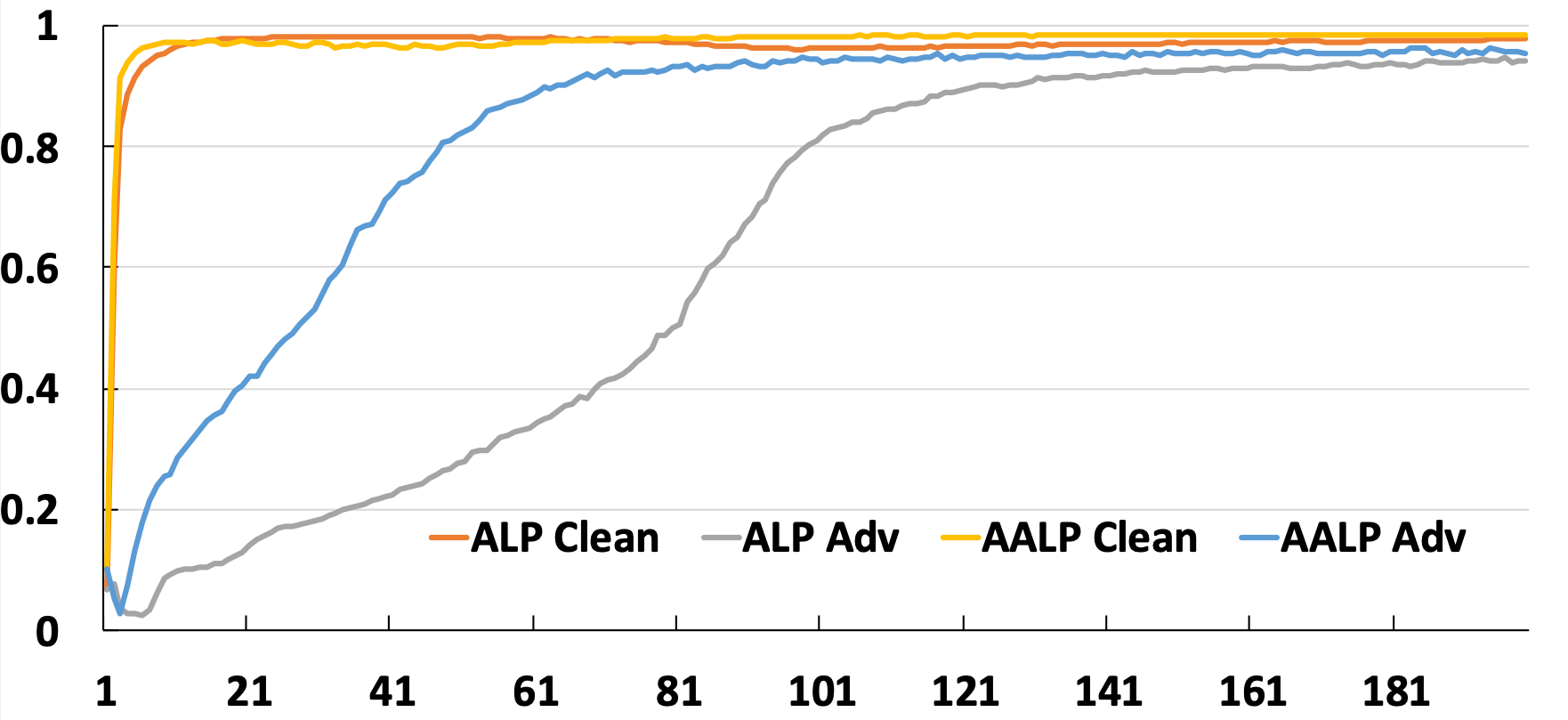}
  \end{center}
    \caption{ALP and AALP training methods convergence speed comparison.}
    \label{speed_comparison}
\end{figure}

It can be seen from Table~\ref{proportion} that under the AALP algorithm, the number of the consistent samples in all three data sets has increased, and the robustness of the model has also been significantly enhanced.

We trained two MNIST models, one with the ALP method applied; the other with the AALP method applied. We record the loss changes of the test set, and the accuracy of clean samples and adversarial samples during the training of the two models.

In Fig.~\ref{speed_comparison}, it can be seen that the AALP algorithm, compared with ALP, has better effects on both convergence and defense.

At the same time, the proposed algorithm solves the problem of ALP loss constraints on different samples; it also reduces the side effects of ALP loss during the training process, accelerates the entire training process, and greatly improves the efficiency of ALP training.

\section{Conclusions}

This work analyzes problems qualitatively and quantitatively in the ALP method: (1) the robust model should reduce the high-contribution features (2) the ALP training target cannot fit all samples. Therefore, we proposed the AALP method, improved the ALP method from the training process and training target, and analyzed the AALP method qualitatively and quantitatively. In the future, we plan to continue to explore the reasons behind the relationship between feature contributions and robustness, and better apply this method to adversarial defense or other computer vision tasks.

\bibliographystyle{ACM-Reference-Format}
\bibliography{sample-base}


\begin{thebibliography}{35}


\ifx \showCODEN    \undefined \def \showCODEN     #1{\unskip}     \fi
\ifx \showDOI      \undefined \def \showDOI       #1{#1}\fi
\ifx \showISBNx    \undefined \def \showISBNx     #1{\unskip}     \fi
\ifx \showISBNxiii \undefined \def \showISBNxiii  #1{\unskip}     \fi
\ifx \showISSN     \undefined \def \showISSN      #1{\unskip}     \fi
\ifx \showLCCN     \undefined \def \showLCCN      #1{\unskip}     \fi
\ifx \shownote     \undefined \def \shownote      #1{#1}          \fi
\ifx \showarticletitle \undefined \def \showarticletitle #1{#1}   \fi
\ifx \showURL      \undefined \def \showURL       {\relax}        \fi
\providecommand\bibfield[2]{#2}
\providecommand\bibinfo[2]{#2}
\providecommand\natexlab[1]{#1}
\providecommand\showeprint[2][]{arXiv:#2}

\bibitem[\protect\citeauthoryear{{Amini} and {Ghaemmaghami}}{{Amini} and
  {Ghaemmaghami}}{2020}]%
        {8970483}
\bibfield{author}{\bibinfo{person}{S. {Amini}} {and} \bibinfo{person}{S.
  {Ghaemmaghami}}.} \bibinfo{year}{2020}\natexlab{}.
\newblock \showarticletitle{Towards Improving Robustness of Deep Neural
  Networks to Adversarial Perturbations}.
\newblock \bibinfo{journal}{\emph{IEEE Transactions on Multimedia}}
  (\bibinfo{year}{2020}), \bibinfo{pages}{1--1}.
\newblock


\bibitem[\protect\citeauthoryear{Athalye, Carlini, and Wagner}{Athalye
  et~al\mbox{.}}{2018}]%
        {athalye2018obfuscated}
\bibfield{author}{\bibinfo{person}{Anish Athalye}, \bibinfo{person}{Nicholas
  Carlini}, {and} \bibinfo{person}{David~A. Wagner}.}
  \bibinfo{year}{2018}\natexlab{}.
\newblock \showarticletitle{Obfuscated Gradients Give a False Sense of
  Security: Circumventing Defenses to Adversarial Examples}. In
  \bibinfo{booktitle}{\emph{Proceedings of the 35th International Conference on
  Machine Learning, {ICML} 2018, Stockholmsm{\"{a}}ssan, Stockholm, Sweden,
  July 10-15, 2018}}. \bibinfo{pages}{274--283}.
\newblock


\bibitem[\protect\citeauthoryear{Carlini and Wagner}{Carlini and
  Wagner}{2017}]%
        {nicholas2017towards}
\bibfield{author}{\bibinfo{person}{Nicholas Carlini} {and}
  \bibinfo{person}{David~A. Wagner}.} \bibinfo{year}{2017}\natexlab{}.
\newblock \showarticletitle{Towards Evaluating the Robustness of Neural
  Networks}. In \bibinfo{booktitle}{\emph{2017 {IEEE} Symposium on Security and
  Privacy, {SP} 2017, San Jose, CA, USA, May 22-26, 2017}}.
  \bibinfo{pages}{39--57}.
\newblock
\urldef\tempurl%
\url{https://doi.org/10.1109/SP.2017.49}
\showDOI{\tempurl}


\bibitem[\protect\citeauthoryear{Carlini and Wagner}{Carlini and
  Wagner}{2018}]%
        {adv_voice}
\bibfield{author}{\bibinfo{person}{Nicholas Carlini} {and}
  \bibinfo{person}{David~A. Wagner}.} \bibinfo{year}{2018}\natexlab{}.
\newblock \showarticletitle{Audio Adversarial Examples: Targeted Attacks on
  Speech-to-Text}. In \bibinfo{booktitle}{\emph{2018 {IEEE} Security and
  Privacy Workshops, {SP} Workshops 2018, San Francisco, CA, USA, May 24,
  2018}}.
\newblock


\bibitem[\protect\citeauthoryear{Choe and Shim}{Choe and Shim}{2019}]%
        {att_dropout}
\bibfield{author}{\bibinfo{person}{Junsuk Choe} {and} \bibinfo{person}{Hyunjung
  Shim}.} \bibinfo{year}{2019}\natexlab{}.
\newblock \showarticletitle{Attention-Based Dropout Layer for Weakly Supervised
  Object Localization}. In \bibinfo{booktitle}{\emph{{IEEE} Conference on
  Computer Vision and Pattern Recognition, {CVPR} 2019, Long Beach, CA, USA,
  June 16-20, 2019}}.
\newblock


\bibitem[\protect\citeauthoryear{Dhillon, Azizzadenesheli, Lipton, Bernstein,
  Kossaifi, Khanna, and Anandkumar}{Dhillon et~al\mbox{.}}{2018}]%
        {dhillon2018stochastic}
\bibfield{author}{\bibinfo{person}{Guneet~S. Dhillon}, \bibinfo{person}{Kamyar
  Azizzadenesheli}, \bibinfo{person}{Zachary~C. Lipton},
  \bibinfo{person}{Jeremy Bernstein}, \bibinfo{person}{Jean Kossaifi},
  \bibinfo{person}{Aran Khanna}, {and} \bibinfo{person}{Anima Anandkumar}.}
  \bibinfo{year}{2018}\natexlab{}.
\newblock \showarticletitle{Stochastic Activation Pruning for Robust
  Adversarial Defense}.
\newblock \bibinfo{journal}{\emph{CoRR}}  \bibinfo{volume}{abs/1803.01442}
  (\bibinfo{year}{2018}).
\newblock
\showeprint[arxiv]{1803.01442}


\bibitem[\protect\citeauthoryear{{Du}, {Fang}, {Yi}, {Xu}, {Cheng}, and
  {Tao}}{{Du} et~al\mbox{.}}{2019}]%
        {8576563}
\bibfield{author}{\bibinfo{person}{Y. {Du}}, \bibinfo{person}{M. {Fang}},
  \bibinfo{person}{J. {Yi}}, \bibinfo{person}{C. {Xu}}, \bibinfo{person}{J.
  {Cheng}}, {and} \bibinfo{person}{D. {Tao}}.} \bibinfo{year}{2019}\natexlab{}.
\newblock \showarticletitle{Enhancing the Robustness of Neural Collaborative
  Filtering Systems Under Malicious Attacks}.
\newblock \bibinfo{journal}{\emph{IEEE Transactions on Multimedia}}
  \bibinfo{volume}{21}, \bibinfo{number}{3} (\bibinfo{date}{March}
  \bibinfo{year}{2019}), \bibinfo{pages}{555--565}.
\newblock
\showISSN{1941-0077}
\urldef\tempurl%
\url{https://doi.org/10.1109/TMM.2018.2887018}
\showDOI{\tempurl}


\bibitem[\protect\citeauthoryear{Ebrahimi, Rao, Lowd, and Dou}{Ebrahimi
  et~al\mbox{.}}{2018}]%
        {adv_text}
\bibfield{author}{\bibinfo{person}{Javid Ebrahimi}, \bibinfo{person}{Anyi Rao},
  \bibinfo{person}{Daniel Lowd}, {and} \bibinfo{person}{Dejing Dou}.}
  \bibinfo{year}{2018}\natexlab{}.
\newblock \showarticletitle{HotFlip: White-Box Adversarial Examples for Text
  Classification}. In \bibinfo{booktitle}{\emph{Proceedings of the 56th Annual
  Meeting of the Association for Computational Linguistics, {ACL} 2018,
  Melbourne, Australia, July 15-20, 2018, Volume 2: Short Papers}}.
\newblock


\bibitem[\protect\citeauthoryear{Feinman, Curtin, Shintre, and Gardner}{Feinman
  et~al\mbox{.}}{2017}]%
        {detecting}
\bibfield{author}{\bibinfo{person}{Reuben Feinman}, \bibinfo{person}{Ryan~R.
  Curtin}, \bibinfo{person}{Saurabh Shintre}, {and} \bibinfo{person}{Andrew~B.
  Gardner}.} \bibinfo{year}{2017}\natexlab{}.
\newblock \showarticletitle{Detecting Adversarial Samples from Artifacts}.
\newblock \bibinfo{journal}{\emph{CoRR}}  \bibinfo{volume}{abs/1703.00410}
  (\bibinfo{year}{2017}).
\newblock


\bibitem[\protect\citeauthoryear{Goodfellow, Shlens, and Szegedy}{Goodfellow
  et~al\mbox{.}}{2015}]%
        {Ian2017Explaining}
\bibfield{author}{\bibinfo{person}{Ian Goodfellow}, \bibinfo{person}{Jonathon
  Shlens}, {and} \bibinfo{person}{Christian Szegedy}.}
  \bibinfo{year}{2015}\natexlab{}.
\newblock \showarticletitle{Explaining and Harnessing Adversarial Examples}. In
  \bibinfo{booktitle}{\emph{International Conference on Learning
  Representations}}.
\newblock


\bibitem[\protect\citeauthoryear{Guo, Rana, Ciss{\'{e}}, and van~der
  Maaten}{Guo et~al\mbox{.}}{2017}]%
        {guo2017countering}
\bibfield{author}{\bibinfo{person}{Chuan Guo}, \bibinfo{person}{Mayank Rana},
  \bibinfo{person}{Moustapha Ciss{\'{e}}}, {and} \bibinfo{person}{Laurens
  van~der Maaten}.} \bibinfo{year}{2017}\natexlab{}.
\newblock \showarticletitle{Countering Adversarial Images using Input
  Transformations}.
\newblock \bibinfo{journal}{\emph{CoRR}}  \bibinfo{volume}{abs/1711.00117}
  (\bibinfo{year}{2017}).
\newblock
\showeprint[arxiv]{1711.00117}


\bibitem[\protect\citeauthoryear{He, Zhang, Ren, and Sun}{He
  et~al\mbox{.}}{2016}]%
        {resnet}
\bibfield{author}{\bibinfo{person}{Kaiming He}, \bibinfo{person}{Xiangyu
  Zhang}, \bibinfo{person}{Shaoqing Ren}, {and} \bibinfo{person}{Jian Sun}.}
  \bibinfo{year}{2016}\natexlab{}.
\newblock \showarticletitle{Deep Residual Learning for Image Recognition}. In
  \bibinfo{booktitle}{\emph{2016 {IEEE} Conference on Computer Vision and
  Pattern Recognition, {CVPR} 2016, Las Vegas, NV, USA, June 27-30, 2016}}.
\newblock


\bibitem[\protect\citeauthoryear{Kannan, Kurakin, and Goodfellow}{Kannan
  et~al\mbox{.}}{2018}]%
        {alp}
\bibfield{author}{\bibinfo{person}{Harini Kannan}, \bibinfo{person}{Alexey
  Kurakin}, {and} \bibinfo{person}{Ian~J. Goodfellow}.}
  \bibinfo{year}{2018}\natexlab{}.
\newblock \showarticletitle{Adversarial Logit Pairing}.
\newblock \bibinfo{journal}{\emph{CoRR}}  \bibinfo{volume}{abs/1803.06373}
  (\bibinfo{year}{2018}).
\newblock


\bibitem[\protect\citeauthoryear{Kingma and Ba}{Kingma and Ba}{2015}]%
        {adam}
\bibfield{author}{\bibinfo{person}{Diederik~P. Kingma} {and}
  \bibinfo{person}{Jimmy Ba}.} \bibinfo{year}{2015}\natexlab{}.
\newblock \showarticletitle{Adam: {A} Method for Stochastic Optimization}. In
  \bibinfo{booktitle}{\emph{3rd International Conference on Learning
  Representations, {ICLR} 2015, San Diego, CA, USA, May 7-9, 2015, Conference
  Track Proceedings}}.
\newblock


\bibitem[\protect\citeauthoryear{Krizhevsky and Hinton}{Krizhevsky and
  Hinton}{2009}]%
        {cifarkrizhevsky2009learning}
\bibfield{author}{\bibinfo{person}{Alex Krizhevsky} {and}
  \bibinfo{person}{Geoffrey Hinton}.} \bibinfo{year}{2009}\natexlab{}.
\newblock \bibinfo{booktitle}{\emph{Learning multiple layers of features from
  tiny images}}.
\newblock \bibinfo{type}{{T}echnical {R}eport}.
  \bibinfo{institution}{Citeseer}.
\newblock


\bibitem[\protect\citeauthoryear{Kurakin, Goodfellow, and Bengio}{Kurakin
  et~al\mbox{.}}{2016}]%
        {kurakin2016adversarial}
\bibfield{author}{\bibinfo{person}{Alexey Kurakin}, \bibinfo{person}{Ian~J.
  Goodfellow}, {and} \bibinfo{person}{Samy Bengio}.}
  \bibinfo{year}{2016}\natexlab{}.
\newblock \showarticletitle{Adversarial examples in the physical world}.
\newblock   \bibinfo{volume}{abs/1607.02533} (\bibinfo{year}{2016}).
\newblock


\bibitem[\protect\citeauthoryear{LeCun, Haffner, Bottou, and Bengio}{LeCun
  et~al\mbox{.}}{1999}]%
        {lenet}
\bibfield{author}{\bibinfo{person}{Yann LeCun}, \bibinfo{person}{Patrick
  Haffner}, \bibinfo{person}{L{\'{e}}on Bottou}, {and} \bibinfo{person}{Yoshua
  Bengio}.} \bibinfo{year}{1999}\natexlab{}.
\newblock \showarticletitle{Object Recognition with Gradient-Based Learning}.
  In \bibinfo{booktitle}{\emph{Shape, Contour and Grouping in Computer
  Vision}}.
\newblock


\bibitem[\protect\citeauthoryear{Liao, Liang, Dong, Pang, Hu, and Zhu}{Liao
  et~al\mbox{.}}{2018a}]%
        {Fangzhouy2017Defense}
\bibfield{author}{\bibinfo{person}{Fangzhou Liao}, \bibinfo{person}{Ming
  Liang}, \bibinfo{person}{Yinpeng Dong}, \bibinfo{person}{Tianyu Pang},
  \bibinfo{person}{Xiaolin Hu}, {and} \bibinfo{person}{Jun Zhu}.}
  \bibinfo{year}{2018}\natexlab{a}.
\newblock \showarticletitle{Defense Against Adversarial Attacks Using
  High-Level Representation Guided Denoiser}. In \bibinfo{booktitle}{\emph{2018
  {IEEE} Conference on Computer Vision and Pattern Recognition, {CVPR} 2018,
  Salt Lake City, UT, USA, June 18-22, 2018}}. \bibinfo{pages}{1778--1787}.
\newblock
\urldef\tempurl%
\url{https://doi.org/10.1109/CVPR.2018.00191}
\showDOI{\tempurl}


\bibitem[\protect\citeauthoryear{Liao, Liang, Dong, Pang, Hu, and Zhu}{Liao
  et~al\mbox{.}}{2018b}]%
        {liao2017defense}
\bibfield{author}{\bibinfo{person}{Fangzhou Liao}, \bibinfo{person}{Ming
  Liang}, \bibinfo{person}{Yinpeng Dong}, \bibinfo{person}{Tianyu Pang},
  \bibinfo{person}{Xiaolin Hu}, {and} \bibinfo{person}{Jun Zhu}.}
  \bibinfo{year}{2018}\natexlab{b}.
\newblock \showarticletitle{Defense Against Adversarial Attacks Using
  High-Level Representation Guided Denoiser}. In \bibinfo{booktitle}{\emph{2018
  {IEEE} Conference on Computer Vision and Pattern Recognition, {CVPR} 2018,
  Salt Lake City, UT, USA, June 18-22, 2018}}. \bibinfo{pages}{1778--1787}.
\newblock
\urldef\tempurl%
\url{https://doi.org/10.1109/CVPR.2018.00191}
\showDOI{\tempurl}


\bibitem[\protect\citeauthoryear{Liu and J{\'{a}}J{\'{a}}}{Liu and
  J{\'{a}}J{\'{a}}}{2018}]%
        {liu2018feature}
\bibfield{author}{\bibinfo{person}{Chihuang Liu} {and} \bibinfo{person}{Joseph
  J{\'{a}}J{\'{a}}}.} \bibinfo{year}{2018}\natexlab{}.
\newblock \showarticletitle{Feature prioritization and regularization improve
  standard accuracy and adversarial robustness}.
\newblock \bibinfo{journal}{\emph{CoRR}}  \bibinfo{volume}{abs/1810.02424}
  (\bibinfo{year}{2018}).
\newblock
\showeprint[arxiv]{1810.02424}


\bibitem[\protect\citeauthoryear{Madry, Makelov, Schmidt, Tsipras, and
  Vladu}{Madry et~al\mbox{.}}{2017}]%
        {madry2017towards}
\bibfield{author}{\bibinfo{person}{Aleksander Madry},
  \bibinfo{person}{Aleksandar Makelov}, \bibinfo{person}{Ludwig Schmidt},
  \bibinfo{person}{Dimitris Tsipras}, {and} \bibinfo{person}{Adrian Vladu}.}
  \bibinfo{year}{2017}\natexlab{}.
\newblock \showarticletitle{Towards Deep Learning Models Resistant to
  Adversarial Attacks}.
\newblock   \bibinfo{volume}{abs/1706.06083} (\bibinfo{year}{2017}).
\newblock


\bibitem[\protect\citeauthoryear{Moosavi{-}Dezfooli, Fawzi, and
  Frossard}{Moosavi{-}Dezfooli et~al\mbox{.}}{2016}]%
        {Seyed-Mohsen2016Deepfool}
\bibfield{author}{\bibinfo{person}{Seyed{-}Mohsen Moosavi{-}Dezfooli},
  \bibinfo{person}{Alhussein Fawzi}, {and} \bibinfo{person}{Pascal Frossard}.}
  \bibinfo{year}{2016}\natexlab{}.
\newblock \showarticletitle{DeepFool: {A} Simple and Accurate Method to Fool
  Deep Neural Networks}. In \bibinfo{booktitle}{\emph{2016 {IEEE} Conference on
  Computer Vision and Pattern Recognition, {CVPR} 2016, Las Vegas, NV, USA,
  June 27-30, 2016}}. \bibinfo{pages}{2574--2582}.
\newblock
\urldef\tempurl%
\url{https://doi.org/10.1109/CVPR.2016.282}
\showDOI{\tempurl}


\bibitem[\protect\citeauthoryear{Netzer, Wang, Coates, Bissacco, Wu, and
  Ng}{Netzer et~al\mbox{.}}{2011}]%
        {SVHN}
\bibfield{author}{\bibinfo{person}{Yuval Netzer}, \bibinfo{person}{Tao Wang},
  \bibinfo{person}{Adam Coates}, \bibinfo{person}{Alessandro Bissacco},
  \bibinfo{person}{Bo Wu}, {and} \bibinfo{person}{Andrew~Y. Ng}.}
  \bibinfo{year}{2011}\natexlab{}.
\newblock \showarticletitle{Reading Digits in Natural Images with Unsupervised
  Feature Learning}. In \bibinfo{booktitle}{\emph{NIPS Workshop on Deep
  Learning and Unsupervised Feature Learning 2011}}.
\newblock


\bibitem[\protect\citeauthoryear{Papernot, McDaniel, Wu, Jha, and
  Swami}{Papernot et~al\mbox{.}}{2016}]%
        {papernot2016distillation}
\bibfield{author}{\bibinfo{person}{Nicolas Papernot},
  \bibinfo{person}{Patrick~D. McDaniel}, \bibinfo{person}{Xi Wu},
  \bibinfo{person}{Somesh Jha}, {and} \bibinfo{person}{Ananthram Swami}.}
  \bibinfo{year}{2016}\natexlab{}.
\newblock \showarticletitle{Distillation as a Defense to Adversarial
  Perturbations Against Deep Neural Networks}. In
  \bibinfo{booktitle}{\emph{{IEEE} Symposium on Security and Privacy, {SP}
  2016, San Jose, CA, USA, May 22-26, 2016}}. \bibinfo{pages}{582--597}.
\newblock
\urldef\tempurl%
\url{https://doi.org/10.1109/SP.2016.41}
\showDOI{\tempurl}


\bibitem[\protect\citeauthoryear{Selvaraju, Cogswell, Das, Vedantam, Parikh,
  and Batra}{Selvaraju et~al\mbox{.}}{2017}]%
        {selvaraju2017grad}
\bibfield{author}{\bibinfo{person}{Ramprasaath~R. Selvaraju},
  \bibinfo{person}{Michael Cogswell}, \bibinfo{person}{Abhishek Das},
  \bibinfo{person}{Ramakrishna Vedantam}, \bibinfo{person}{Devi Parikh}, {and}
  \bibinfo{person}{Dhruv Batra}.} \bibinfo{year}{2017}\natexlab{}.
\newblock \showarticletitle{Grad-CAM: Visual Explanations from Deep Networks
  via Gradient-Based Localization}. In \bibinfo{booktitle}{\emph{{IEEE}
  International Conference on Computer Vision, {ICCV} 2017, Venice, Italy,
  October 22-29, 2017}}. \bibinfo{pages}{618--626}.
\newblock
\urldef\tempurl%
\url{https://doi.org/10.1109/ICCV.2017.74}
\showDOI{\tempurl}


\bibitem[\protect\citeauthoryear{Sharif, Bhagavatula, Bauer, and Reiter}{Sharif
  et~al\mbox{.}}{2018}]%
        {adv_face}
\bibfield{author}{\bibinfo{person}{Mahmood Sharif}, \bibinfo{person}{Sruti
  Bhagavatula}, \bibinfo{person}{Lujo Bauer}, {and} \bibinfo{person}{Michael~K.
  Reiter}.} \bibinfo{year}{2018}\natexlab{}.
\newblock \showarticletitle{Adversarial Generative Nets: Neural Network Attacks
  on State-of-the-Art Face Recognition}.
\newblock \bibinfo{journal}{\emph{CoRR}}  \bibinfo{volume}{abs/1801.00349}
  (\bibinfo{year}{2018}).
\newblock


\bibitem[\protect\citeauthoryear{Shen, Jin, Gao, and Zhang}{Shen
  et~al\mbox{.}}{2017}]%
        {shen2017ape}
\bibfield{author}{\bibinfo{person}{Shiwei Shen}, \bibinfo{person}{Guoqing Jin},
  \bibinfo{person}{Ke Gao}, {and} \bibinfo{person}{Yongdong Zhang}.}
  \bibinfo{year}{2017}\natexlab{}.
\newblock \showarticletitle{Ape-gan: Adversarial perturbation elimination with
  gan}.
\newblock \bibinfo{journal}{\emph{ICLR Submission, available on OpenReview}}
  \bibinfo{volume}{4} (\bibinfo{year}{2017}).
\newblock


\bibitem[\protect\citeauthoryear{Song, Kim, Nowozin, Ermon, and Kushman}{Song
  et~al\mbox{.}}{2017}]%
        {song2017pixeldefend}
\bibfield{author}{\bibinfo{person}{Yang Song}, \bibinfo{person}{Taesup Kim},
  \bibinfo{person}{Sebastian Nowozin}, \bibinfo{person}{Stefano Ermon}, {and}
  \bibinfo{person}{Nate Kushman}.} \bibinfo{year}{2017}\natexlab{}.
\newblock \showarticletitle{PixelDefend: Leveraging Generative Models to
  Understand and Defend against Adversarial Examples}.
\newblock \bibinfo{journal}{\emph{CoRR}}  \bibinfo{volume}{abs/1710.10766}
  (\bibinfo{year}{2017}).
\newblock
\showeprint[arxiv]{1710.10766}


\bibitem[\protect\citeauthoryear{Srivastava, Hinton, Krizhevsky, Sutskever, and
  Salakhutdinov}{Srivastava et~al\mbox{.}}{2014}]%
        {dropout}
\bibfield{author}{\bibinfo{person}{Nitish Srivastava},
  \bibinfo{person}{Geoffrey~E. Hinton}, \bibinfo{person}{Alex Krizhevsky},
  \bibinfo{person}{Ilya Sutskever}, {and} \bibinfo{person}{Ruslan
  Salakhutdinov}.} \bibinfo{year}{2014}\natexlab{}.
\newblock \showarticletitle{Dropout: a simple way to prevent neural networks
  from overfitting}.
\newblock \bibinfo{journal}{\emph{J. Mach. Learn. Res.}}
  (\bibinfo{year}{2014}).
\newblock


\bibitem[\protect\citeauthoryear{{Su}, {Fang}, {Wang}, {Mehrotra}, {Begen},
  {Ye}, and {Cavallaro}}{{Su} et~al\mbox{.}}{2019}]%
        {8649865}
\bibfield{author}{\bibinfo{person}{Z. {Su}}, \bibinfo{person}{Q. {Fang}},
  \bibinfo{person}{H. {Wang}}, \bibinfo{person}{S. {Mehrotra}},
  \bibinfo{person}{A.~C. {Begen}}, \bibinfo{person}{Q. {Ye}}, {and}
  \bibinfo{person}{A. {Cavallaro}}.} \bibinfo{year}{2019}\natexlab{}.
\newblock \showarticletitle{Guest Editorial Trustworthiness in Social
  Multimedia Analytics and Delivery}.
\newblock \bibinfo{journal}{\emph{IEEE Transactions on Multimedia}}
  \bibinfo{volume}{21}, \bibinfo{number}{3} (\bibinfo{year}{2019}),
  \bibinfo{pages}{537--538}.
\newblock


\bibitem[\protect\citeauthoryear{Szegedy, Zaremba, Sutskever, Bruna, Erhan,
  Goodfellow, and Fergus}{Szegedy et~al\mbox{.}}{2014}]%
        {szegedy2013intriguing}
\bibfield{author}{\bibinfo{person}{Christian Szegedy},
  \bibinfo{person}{Wojciech Zaremba}, \bibinfo{person}{Ilya Sutskever},
  \bibinfo{person}{Joan Bruna}, \bibinfo{person}{Dumitru Erhan},
  \bibinfo{person}{Ian Goodfellow}, {and} \bibinfo{person}{Rob Fergus}.}
  \bibinfo{year}{2014}\natexlab{}.
\newblock \showarticletitle{Intriguing properties of neural networks}. In
  \bibinfo{booktitle}{\emph{International Conference on Learning
  Representations}}.
\newblock


\bibitem[\protect\citeauthoryear{Tram{\`{e}}r, Kurakin, Papernot, Boneh, and
  McDaniel}{Tram{\`{e}}r et~al\mbox{.}}{2017}]%
        {tramer2017ensemble}
\bibfield{author}{\bibinfo{person}{Florian Tram{\`{e}}r},
  \bibinfo{person}{Alexey Kurakin}, \bibinfo{person}{Nicolas Papernot},
  \bibinfo{person}{Dan Boneh}, {and} \bibinfo{person}{Patrick~D. McDaniel}.}
  \bibinfo{year}{2017}\natexlab{}.
\newblock \showarticletitle{Ensemble Adversarial Training: Attacks and
  Defenses}.
\newblock   \bibinfo{volume}{abs/1705.07204} (\bibinfo{year}{2017}).
\newblock


\bibitem[\protect\citeauthoryear{{Wang}, {Su}, {Zhang}, and {Hu}}{{Wang}
  et~al\mbox{.}}{2019}]%
        {8884184}
\bibfield{author}{\bibinfo{person}{Y. {Wang}}, \bibinfo{person}{H. {Su}},
  \bibinfo{person}{B. {Zhang}}, {and} \bibinfo{person}{X. {Hu}}.}
  \bibinfo{year}{2019}\natexlab{}.
\newblock \showarticletitle{Learning Reliable Visual Saliency for Model
  Explanations}.
\newblock \bibinfo{journal}{\emph{IEEE Transactions on Multimedia}}
  (\bibinfo{year}{2019}), \bibinfo{pages}{1--1}.
\newblock


\bibitem[\protect\citeauthoryear{Xie, Wang, Zhang, Zhou, Xie, and Yuille}{Xie
  et~al\mbox{.}}{2017}]%
        {advobdetection}
\bibfield{author}{\bibinfo{person}{Cihang Xie}, \bibinfo{person}{Jianyu Wang},
  \bibinfo{person}{Zhishuai Zhang}, \bibinfo{person}{Yuyin Zhou},
  \bibinfo{person}{Lingxi Xie}, {and} \bibinfo{person}{Alan~L. Yuille}.}
  \bibinfo{year}{2017}\natexlab{}.
\newblock \showarticletitle{Adversarial Examples for Semantic Segmentation and
  Object Detection}. In \bibinfo{booktitle}{\emph{{IEEE} International
  Conference on Computer Vision, {ICCV} 2017, Venice, Italy, October 22-29,
  2017}}.
\newblock


\bibitem[\protect\citeauthoryear{Zhang, Yu, Jiao, Xing, Ghaoui, and
  Jordan}{Zhang et~al\mbox{.}}{2019}]%
        {ZhangYJXGJ19}
\bibfield{author}{\bibinfo{person}{Hongyang Zhang}, \bibinfo{person}{Yaodong
  Yu}, \bibinfo{person}{Jiantao Jiao}, \bibinfo{person}{Eric~P. Xing},
  \bibinfo{person}{Laurent~El Ghaoui}, {and} \bibinfo{person}{Michael~I.
  Jordan}.} \bibinfo{year}{2019}\natexlab{}.
\newblock \showarticletitle{Theoretically Principled Trade-off between
  Robustness and Accuracy}. In \bibinfo{booktitle}{\emph{Proceedings of the
  36th International Conference on Machine Learning, {ICML} 2019, 9-15 June
  2019, Long Beach, California, {USA}}}. \bibinfo{pages}{7472--7482}.
\newblock


\end{thebibliography}

\end{document}